\documentclass[10pt,twocolumn,letterpaper]{article}

\usepackage{iccv} 
\usepackage{graphicx}
\usepackage{subcaption}
\usepackage{multirow}

\usepackage{xspace}
\newcommand{\pname}{RoSe\xspace}

\definecolor{iccvblue}{rgb}{0.21,0.49,0.74}
\usepackage[pagebackref,breaklinks,colorlinks,allcolors=iccvblue]{hyperref}

\def\paperID{6948} 
\def\confName{ICCV}
\def\confYear{2025}

\title{Robust Low-light Scene Restoration via Illumination Transition}
\author{
Ze Li$^{1,2}$ \quad
Feng Zhang$^{3,2}$ \quad
Xiatian Zhu$^{4}$ \quad
Meng Zhang$^{1,2}$ \quad
Yanghong Zhou$^{2}$ \quad
P. Y. Mok$^{1}$\thanks{Corresponding author: tracy.mok@ust.hk} \\
$^{1}$The Hong Kong University of Science and Technology, Hong Kong SAR \\
$^{2}$The Hong Kong Polytechnic University, Hong Kong SAR \\
$^{3}$Nanjing University of Posts and Telecommunications, Nanjing, China\\
$^{4}$University of Surrey, Guildford, United Kingdom  \\
}

\begin{document}
\maketitle

\begin{abstract}
Synthesizing normal-light novel views from low-light multiview images is an important yet challenging task, given the low visibility and high ISO noise present in the input images.
Existing low-light enhancement methods often struggle to effectively preprocess such low-light inputs, as they fail to consider correlations among multiple views.
Although other state-of-the-art methods have introduced illumination-related components offering alternative solutions to the problem, they often result in drawbacks such as color distortions and artifacts, and they provide limited denoising effectiveness.
In this paper, we propose a novel \textbf{R}obust L\textbf{o}w-light \textbf{S}cene R\textbf{e}storation framework (\textbf{\pname}), which enables effective synthesis of novel views in normal lighting conditions from low-light multiview image inputs, by formulating the task as an illuminance transition estimation problem in 3D space, conceptualizing it as a specialized rendering task.
This multiview-consistent illuminance transition field establishes a robust connection between low-light and normal-light conditions. By further exploiting the inherent low-rank property of illumination to constrain the transition representation, we achieve more effective denoising without complex 2D techniques or explicit noise modeling.
To implement \pname, we design a concise dual-branch architecture and introduce a low-rank denoising module.
Experiments demonstrate that \pname significantly outperforms state-of-the-art models in both rendering quality and multiview consistency on standard benchmarks.
The codes and data are available at \url{https://pegasus2004.github.io/RoSe}.
\end{abstract}

\section{Introduction}
The synthesis of a normal-light 3D scene from low-light multiview images \cite{cui2024aleth,wang2023lighting} remains a valuable yet challenging task, as in many real-world applications 
where well-lit data acquisition is often not feasible.
Recent advancements in techniques such as Neural Radiance Fields (NeRFs) \cite{mildenhall2021nerf} have demonstrated impressive capabilities in 3D scene representation; however, they struggle in low-light conditions because of the reliance on direct color optimization, which becomes unreliable in cases of low visibility or high ISO noise.
\begin{figure}[t]
    \centering
    \includegraphics[width=0.47\textwidth]{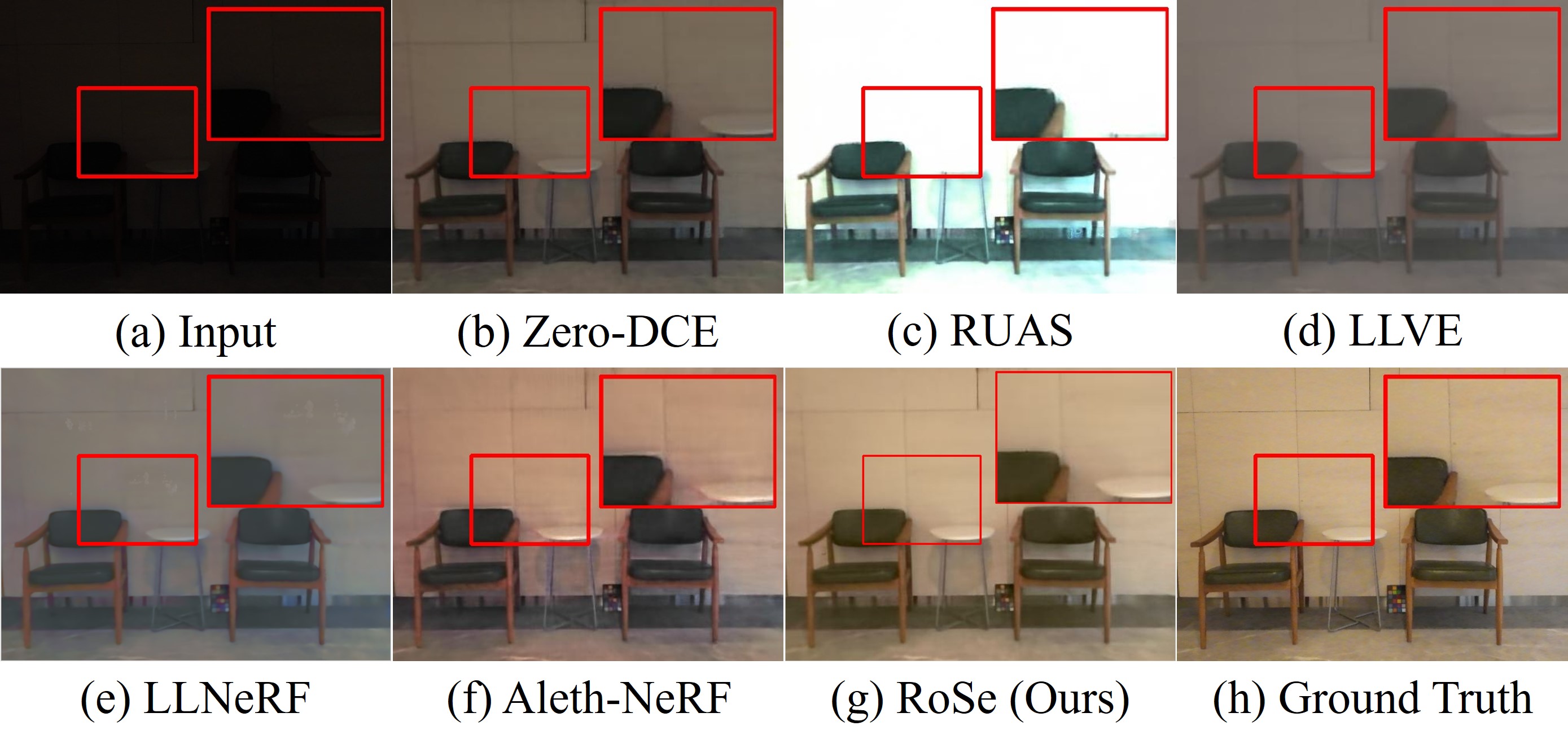}
    \vspace{-3mm}
    \caption{
    Image/video enhancement (Zero-DCE \cite{guo2020zero}+NeRF, RUAS \cite{liu2021retinex} +NeRF, LLVE \cite{zhang2021learning} +NeRF) {\em vs.}
 state-of-the-art models (LLNeRF \cite{wang2023lighting}, Aleth-Nerf \cite{cui2024aleth}) {\em vs.} our \pname.}
    \label{fig:motivation}
\end{figure}

Exsiting low-light enhancement methods \cite{ma2022toward,cui2022you,liu2021retinex,zhang2021learning,guo2020zero} offer a straightforward solution by converting low-light inputs into normal-light images either before or after 3D scene optimization.
However, these methods often result in low-quality novel view images with artifacts and color distortion (\cref{fig:motivation}(b-d)), as these enhancement methods primarily focus on improving individual views independently without considering the multiview relationship.

More recently, new approaches are proposed. For example, 
LLNeRF \cite{wang2023lighting} improves brightness by learning per-view enhancement coefficients for the view-dependent color component of 3D sampling points.
However, since the view-dependent components decoupled from the same sampling point vary across different viewpoints, LLNeRF may introduce conflicting enhancements for that point, leading to artifacts and color distortion in the synthesized images (\cref{fig:motivation}(e)). 
Aleth-NeRF \cite{cui2024aleth} proposes a counter intuitive yet effective approach by assigning non-zero transmittance values to air particles to model light attenuation. 
However, this makes air behave more like a light-absorbing medium such as fog or water. 
When there is a large brightness gap between the low-light observations and the target restored scene, high opacity values may cause the network to misinterpret air as part of the objects and lead to color distortion (\cref{fig:motivation}(f) and \cref{fig:ab-light}).
Moreover, both methods rely solely on NeRF's multiview optimization for implicit denoising.
If noise remains consistent across views, they may misinterpret such noise as part of the scene geometry, leading to artifacts.

In this work, we present \textbf{\pname}, a \textbf{r}obust l\textbf{o}w-light \textbf{s}cene r\textbf{e}storation framework for synthetic normal-light novel views from noisy, low-light multiview inputs. 
Inspired by 2D Retinex theory \cite{rahman2004retinex}, we frame this task as an illuminance transition estimation problem in 3D space and conceptualize it as a specialized rendering task.
Specifically, \pname employs a dual-branch architecture consisting of two key components: a viewer-centered normal-light scene representation that captures the light interactions between objects and viewers, and a world-centered illuminance transition estimation that learns multiview-consistent transformations from low-light to normal-light settings.
The two branches work together to represent the observed low-light scenes, establishing a robust connection between low-light and normal-light conditions.

This design is further enhanced by leveraging the inherent low-rank nature of illumination, which contrasts with the high-rank randomness of noise.
To this end, a low-rank denoising module (LRD) is integrated into the illuminance transition estimation branch to eliminate artifacts and improve the structural integrity of the scene.
As shown in \cref{fig:motivation}(g), \pname can render clear, normal-light novel views from low-light inputs, outperforming the state-of-the-art methods \cite{wang2023lighting, cui2024aleth} and other light-enhancement based methods \cite{cui2022you,liu2021retinex,guo2020zero}.

Our contributions are as follows:
\begin{itemize}
\item We reformulate the low-light scene restoration task as an illuminance transition estimation problem.
\item We further conceptualize this illuminance transition estimation process as a specialized rendering problem, leveraging the low-rank property of illumination for effective denoising.
\item We propose \pname a novel architecture with a robust unsupervised illumination-aware restoration for this task.
\item Experiments show that \pname achieves superior visual quality and multiview consistency compared to state-of-the-art alternatives.
\end{itemize}
\vspace{-1mm}

\section{Related Work}
\label{related}

\subsection{Neural Radiance Field}

Neural Radiance Fields \cite{mildenhall2021nerf} represent continuous 3D scenes by predicting density and color information for each point in 3D space, enabling high-quality novel view synthesis. 
While recent advances have addressed various limitations of NeRF, such as dense view requirements \cite{somraj2023vip,niemeyer2022regnerf,yu2021pixelnerf,jain2021putting} and long training times \cite{fridovich2022plenoxels,reiser2021kilonerf,garbin2021fastnerf}, significant challenges remain in handling adverse lighting conditions. 
Many studies have aimed to improve NeRF’s robustness in adverse lighting environments, such as images with unconstrained exposure \cite{ren2024nerf,kim2024up,sabour2023robustnerf,martin2021nerf,zhang2021nerfactor}, relighting \cite{toschi2023relight,xu2023renerf,rudnev2022nerf}, and underwater scenes \cite{tang2024neural,zhang2023beyond}. 
However, low-light scene restoration presents unique challenges for NeRF due to low visibility and high ISO noise, making accurate geometry inference difficult. 

For 3D scenes with low-light inputs, 
RAW-NeRF \cite{mildenhall2022nerf} renders NeRF directly on noisy raw HDR images, enhanced with an image signal processor (ISP).
However, the requirement of HDR RAW data for training makes it hard to generalize on commonly used sRGB images.
LLNeRF \cite{wang2023lighting} utilizes NeRF's multiview optimization for denoising. It also proposes to decompose and manipulate illumination components along viewing rays to enhance the brightness. However, the view-dependent decomposition suffers from inconsistent illumination factorization across different viewpoints due to varying ray sampling distributions, leading to unstable rendering results.
Aleth-NeRF \cite{cui2024aleth} introduced concealing fields that model light attenuation from normal-light to low-light scenes by assigning non-zero transmittance values to air particles.  
By removing the learned concealing fields, Aleth-NeRF can achieve effective scene illumination restoration.
However, when there is a significant illumination gap between low-light observation and normal-light scenes, Aleth-NeRF assigns high opacity to air particles, leading to the network misjudging them as part of the object.

In this work, we reformulate the task as an illuminance transition estimation problem.
By modeling illuminance transition as a specialized neural field, we establish a physically consistent mapping between low-light and normal-light scene representations, enabling robust novel view synthesis while preserving scene structure.

\subsection{Low-light Enhancement}

In recent years, advances in deep learning have markedly improved low-light image enhancement (LIE) performance. 
Many existing methods \cite{yang2023implicit,xu2022snr,wu2022uretinex,sharma2021nighttime,yang2021sparse} 
rely on paired training data for brightness enhancement, yet capturing paired low-light and normal-light images in real-world settings is impractical. 
To address this, unsupervised GAN-based approaches, such as EnlightenGAN \cite{jiang2021enlightengan}, eliminate the need for paired data but require careful curation of unpaired data. 
Alternatively, zero-reference methods \cite{ma2022toward,li2021learning,liu2021retinex,zhu2020zero,guo2020zero} avoid using ground-truth supervision altogether, relying on encoded priors to guide enhancement.

The Retinex theory \cite{rahman2004retinex} provides a fundamental framework for low-light enhancement by mathematically relating low-light observations to normal-light images, which has inspired many effective solutions.
For instance, Liu et al. \cite{liu2021retinex} use Retinex principles to develop an enhancement model that unrolls illumination estimation and noise removal steps within an optimization framework, augmented by learnable architecture search. Similarly, Ma et al. \cite{ma2022toward} propose a Self-Calibrated Illumination (SCI) learning framework based on Retinex theory, where a self-calibration module estimates reflectance, iteratively refining illumination maps to produce enhanced outputs after illumination is removed.

Despite these advances, existing methods predominantly operate in 2D image space, failing to leverage the inherent 3D geometry of scenes or accommodate multi-view inputs effectively. 
While video enhancement methods \cite{zhang2021learning,lv2018mbllen} have made progress in maintaining temporal consistency between adjacent frames, they primarily focus on enhancing original viewpoints rather than generating coherent 3D scenes with novel views.

Unlike traditional Retinex-based methods confined to 2D applications, we explicitly model an illuminance transition field to bridge the normal-light scene representation and the low-light observed scene. 
This illuminance transition field represents a fundamental shift in how illumination is handled in multiview scenarios, enabling coherent illuminance transition across different viewpoints while preserving scene structure.

\section{Robust Low-light Scene Restoration}

\subsection{Motivation}
In Retinex theory \cite{rahman2004retinex}, 
an image $Z \in \mathbf{R}^{H\times W \times3}$ can be represented as reflectance $R \in \mathbf{R}^{H\times W \times3}$ and illumination $L \in \mathbf{R}^{H\times W \times1}$ as follows:
\begin{equation}
    Z = R \odot L,
\label{eq1}
\end{equation}
where $\odot$ denotes element-wise multiplication.
Many Retinex-based methods \cite{WeiWY018,liu2021retinex,ma2022toward,wu2022uretinex,yi2023diff,cai2023retinexformer} adopt a decomposition-and-enhancement pipeline to restore the normal-light image $Z_{nor}$ from the low-light observations $Z_{low}$. 
However, the decomposition is ill-posed, requiring additional constraints on both $R$ and $L$.
Since the illumination can not be thoroughly decomposed, the enhancement process may result in color distortion, requiring ground truth data for supervision. 

To address these limitations, we introduce an illuminance transition component $I\in \mathbf{R}^{H\times W \times1}$ to establish a direct connection between $Z_{nor}$ and $Z_{low}$, thereby reformulating low-light scene restoration as an illuminance transition estimation problem.
According to the Retinex theory \cite{rahman2004retinex}, reflectance $R$ is an invariant intrinsic property of a scene, i.e., $R_{nor} = R_{low}$.
Given $Z_{nor} = R_{nor} \odot L_{nor}$ and $Z_{low} = R_{low} \odot L_{low}$, this leads to the following formulation:
\begin{equation}
Z_{low} = Z_{nor} \odot I,
\end{equation}
where $ I = \frac{L_{low}}{L_{nor}}.$
This simplified formulation enables direct recovery of normal-light image $Z_{nor}$ from low-light input $Z_{low}$ by estimating the illuminance transition $I$, bypassing the need for two-stage decomposition and enhancement.
As $I$ encodes the relative illumination ratio $\frac{L_{low}}{L_{nor}}$ rather than directly modulating illumination $L$ or reflectance $R$, it inherently preserves the scene’s color properties, mitigating the potential color deviation issue.
Building on this insight, we extend this formulation to 3D space, introducing \textbf{RoSe}, a \textbf{r}obust l\textbf{o}w-light \textbf{s}cene r\textbf{e}storation framework.

\begin{figure*}[t]
    \centering
    \includegraphics[width=1\textwidth]{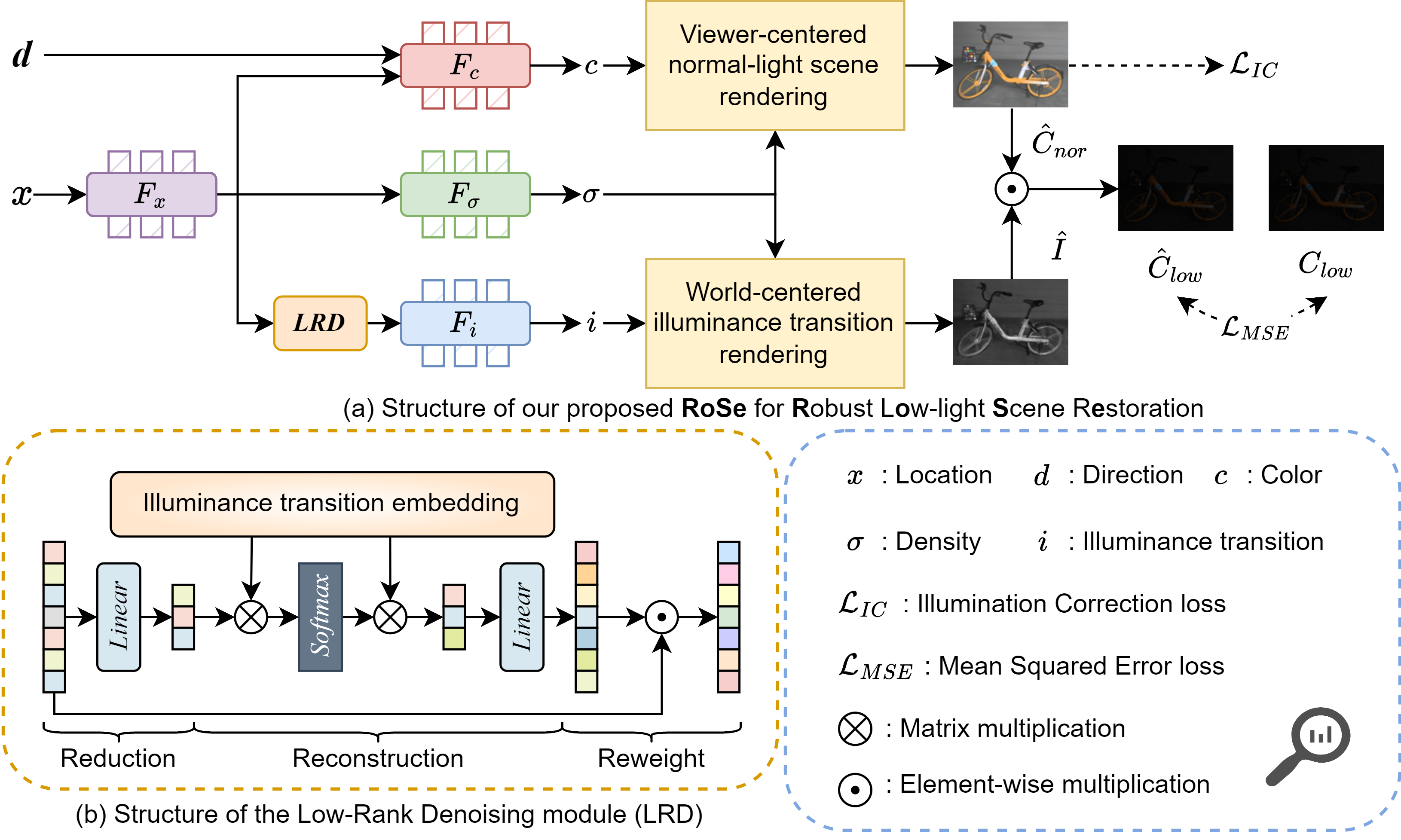}
    \vspace{-3mm}
    \caption{An overview of the proposed RoSe framework. It learns both normal-light scene representation $(\textbf{c}, \sigma)$ and illuminance transition $(i, \sigma)$ from the low-light scene. 
    A low-rank denoising module (LRD) is integrated into the illuminance transition estimation branch, leveraging the low-rank property of illumination for denoising.
    Through volume rendering (\cref{eq6}, \cref{eq8}), it generates illuminance transition value $\hat{I}$ and normal-light RGB values $\hat{C}_{nor}$, which combine to represent observed low-light pixels $\hat{C}_{low}$ (\cref{eq3}). The network is trained using only low-light observed images as supervision through a mean square error loss as used in NeRF, complemented by an unsupervised illumination correction loss.}
    \label{fig:network}
\end{figure*}

\subsection{Architecture}

In 3D space, we treat each pixel in the image $Z$ and the illuminance transition component $I$ as the result of a volume rendering process, where the accumulated radiance and illuminance transition value along each camera ray yield the observed color $\hat{C}$ and illuminance component $\hat{I}$, respectively. 
Based on this view, we conceptualize illuminance transition estimation as a specialized rendering problem, resulting in a concise dual-branch architecture.

As shown in \cref{fig:network} (a), \pname takes 3D location $\textbf{x}$ and view direction $\textbf{d}$ as input to jointly model the normal-light scene representation $(\textbf{c},\sigma)$ and the illuminance transition field $(i,\sigma)$. 
The normal-light scene learning is viewer-centered, capturing the lighting effects between the observer and the object. 
The illuminance transition learning is world-centered, ensuring consistent illuminance transitions across different viewpoints. 
Through volume rendering, it generates normal-light RGB value $\hat{C}_{nor}$ and illumination transition value $\hat{I}$, which together reconstruct the observed low-light pixels $\hat{C}_{low}$:
\begin{equation}
   \hat{C}_{low} = \hat{C}_{nor}\odot \hat{I}.
   \label{eq3}
\end{equation}

Since noise is a 2D phenomenon caused by the imaging sensor, 
we neither model it explicitly in 3D space nor rely on additional 2D denoising techniques.
Instead, we leverage the low-rank property of illumination for noise suppression and introduce a low-rank denoising (LRD) module within our illuminance transition branch (\cref{fig:network} (b)).

The network is trained using only low-light observed images as supervision through a mean square error loss.
Additionally, an unsupervised illumination correction loss ensures that the normal-light scene recovers to the target illumination level. 
The following sections will detail the core components of our framework.

\noindent
{\bf Viewer-centered normal-light scene rendering.}
The normal-light scene representation branch follows the standard NeRF formulation:
\begin{equation}
    F(\textbf{x},\textbf{d}) \rightarrow (\textbf{c},\sigma) ,
\end{equation}
where $\textbf{x}=(x,y,z)$ denotes the 3D spatial coordinates of sample points along a camera ray, and $\textbf{d}=(\theta,\phi)$ represents the viewing direction. $\textbf{c}=(r,g,b)$ and $\sigma$ are predicted color and volume density of each sample point. 
This formulation captures the interaction of light with the scene geometry from a location to the viewer, making it inherently viewer-centered.

For the network structure, we adopt three multilayer perceptrons (MLPs): a location MLP $F_x$, a density MLP $F_{\sigma}$, and a color MLP $F_c$.
These MLPs map the 3D spatial coordinates $\textbf{x}$ and viewing direction $\textbf{d}$ to the corresponding density $\sigma$ and color $\textbf{c}$:
\begin{equation}
     \sigma = F_{\sigma}(F_x(\textbf{x})),\\
     \textbf{c} = F_c(F_x(\textbf{x}),\textit{\textbf{d}}).
\end{equation}
The predicted color and density values of all sample points along a camera ray $\textbf{r}$ are then integrated to render the expected color $\hat{C}_{nor}(\textbf{r})$.
In practice, the integrals are approximated through discrete computation:
\begin{equation}
    \hat{C}_{nor}(\textbf{r}) = \sum_{n=1}^N T_n\, (1-exp(-\sigma_n\delta_n))\, \textbf{c}_n, 
\label{eq6}
\end{equation}
where $T_n = \exp\left(-\sum_{j=1}^{n-1} \sigma_j \delta_j\right)$, $N$ is the total number of sample points and $\delta_n=t_{n+1}-t_n$ is the distance between adjacent samples.

\noindent
{\bf World-centered illuminance transition rendering.}
As normal-light scene representation already captures view-dependent effects, we argue that the illuminance transition should be world-centered. 
When the illuminance transition field remains consistent across different viewpoints, it ensures stable illumination transitions and effectively avoids artifacts. 

To achieve this, we introduce a world-centered illuminance transition estimation branch.
This branch includes an illuminance transition MLP $F_i$ for estimating $i$, which is solely dependent on the 3D spatial location:
\begin{equation}
     i=F_i(F_x(x)).
\end{equation}

Similar to the normal-light scene rendering, we use a discrete approximation to estimate the integral of the predicted illuminance transition value $i$ and obtain the illuminance transition value $\hat{I}(\textbf{r})$  along a camera ray:
\begin{equation}
   \hat{I}(\textbf{r}) = \sum_{n=1}^N  T_n\, (1-exp(-\sigma_n\delta_n))\,i_n, 
\label{eq8}
\end{equation}
where $T_n = \exp\left(-\sum_{j=1}^{n-1} \sigma_j \delta_j\right)$, $N$ is the total number of sample points and $\delta_n=t_{n+1}-t_n$ is the distance between adjacent samples.

\noindent
{\bf Low-rank denoising.}
The spatial smoothness of illumination inherently reflects its low-rank nature, while noise is typically random and high-rank. 
Thus, effective denoising can be achieved by simply constraining this property, without relying on 2D denoising techniques or explicit noise modeling.

As shown in \cref{fig:network} (b), our low-rank denoising module follows a reduction-reconstruction-reweight process to implement low-rank regularization.
In the first stage, the input feature $f_x \in \mathbb{R}^{B\times C}$ is projected into a low-rank subspace via a linear transformation, producing $f_k \in \mathbb{R}^{B\times k}$:
 \begin{equation}
     f_k=Linear(f_x),
\end{equation}
After that, a set of learnable low-rank filters is introduced as the illuminance transition embedding $E \in \mathbb{R}^{k\times M}$. 
Inspired by the attention mechanism \cite{vaswani2017attention}, we then calculate the similarity between $f_k$ and $E$ to reconstruct the most relevant low-rank feature $f_g\in \mathbb{R}^{B\times k}$:
\begin{equation}
     w=Softmax(f_k \otimes E),
     f_g = w \otimes E,
\end{equation}
where $\otimes$ denotes matrix multiplication.

Rather than directly replacing the original feature, $f_g$ serves as a guidance signal. This avoids the limitations of low-rank representations, which may lose spatial details.
The reconstructed feature is expanded back to the original dimensionality and used to reweight the input:
\begin{equation}
     f_g'=Linear(f_g),
     f_i=f'_g \odot f_x.
\end{equation}
By doing so, we can suppress high-rank noise, thereby improving object geometry estimation.

\subsection{Optimization}

During training, the normal-light scene representation is optimized unsupervisedly. Specifically, an illumination correction loss is used to regulate the global illumination level, while a reconstruction loss between low-light observations and reconstructed images is applied to learn the underlying scene geometry.
Details are provided below.

\noindent
{\bf Illumination correction loss.}
We introduce a hyperparameter $e$ to represent the desired illumination level for the normal-light scene and apply an unsupervised illumination correction loss $\mathcal{L}_{IC}$ to the rendered normal-light image, regulating it by minimizing the difference between the average intensity of the rendered image and the desired illumination level $e$:
\begin{equation}
    \mathcal{L}_{IC} = \left\| GAP(\hat{C}_{nor}(\mathbf{r}))- e \right\|^2,
\end{equation}
where $GAP(\cdot)$ denotes global average pooling. 
In our experiments, we set $e$ to 0.45 to achieve balanced performance.

\noindent
{\bf NeRF regression loss.}
As dark pixels with small values would contribute minimally, while bright pixels with larger values would dominate the contribution. We apply an inverse tone curve \cite{brooks2019unprocessing} to rebalance pixel weights before computing the MSE loss:
\begin{equation}
\begin{split}
    \phi(x) &= \frac{1}{2} - \sin\left(\frac{\sin^{-1}(1 - 2x)}{3}\right), \\
    \mathcal{L}_{\text{MSE}} &= \sum_{r}^R \left\|\hat{C}_{low}(\textbf{r} ) - \phi(C_{low}(\textbf{r}) + \varepsilon) \right\|^2,
\end{split}
\end{equation}
where $\varepsilon = 1e-3$, and $R$ is the total number of pixels in the reconstructed image.
Notably, while the inverse tone curve rebalances pixel weights, it also amplifies noise, which necessitates dedicated denoising.

\begin{table*}[h!]
    \centering
    \caption{The quantitative comparison results between ours and existing methods on test scenes with paired normal-light images.
    Metrics: PSNR(P)$\uparrow$,SSIM(S)$\uparrow$, and LPIPS(L)$\downarrow$. \textbf{Bold}: The best result; \underline{underline}: The second best result.}
    \vspace{-3mm}
    \resizebox{1\textwidth}{!}{
    \begin{tabular}{cccccccc}
        \toprule
         \multirow{2}{*}{\textbf{Method}} & \textbf{“buu”} & \textbf{“chair”} & \textbf{“sofa”} & \textbf{“bike”} & \textbf{“shrub”} & \textbf{mean} \\
    \cline{2-7}
    & PSNR/SSIM/LPIPS & PSNR/SSIM/LPIPS & PSNR/SSIM/LPIPS & PSNR/SSIM/LPIPS & PSNR/SSIM/LPIPS & PSNR/SSIM/LPIPS \\
        \midrule
        NeRF \cite{mildenhall2021nerf} & 7.51/0.291/0.448 & 6.04/0.147/0.594 & 6.28/0.210/0.568 & 6.35/0.072/0.623 & 8.03/0.201/0.680 & 6.84/0.150/0.582 \\
        \midrule
        \multicolumn{7}{c}{Image enhancement methods} \\
        \midrule
        LIME \cite{guo2016lime} & 13.92/0.795/0.302 & 11.33/0.698/0.509 & 12.26/0.759/0.444 & 11.26/0.589/0.439 & 14.16/0.424/0.465 & 12.59/0.653/0.432\\
        DUAL \cite{zhang2019dual} & 13.11/0.770/0.305 & 10.36/0.651/0.500 & 11.20/0.713/0.429 & 9.24/0.433/0.470 & 13.97/0.415/0.470 &11.58/0.596/0.435 \\
        RetinexNet \cite{WeiWY018} & 16.19/0.780/0.396 & 16.89/0.756/0.543 & 16.98/0.807/0.577 & 18.00/0.707/0.482 & 14.86/0.284/0.518 & 16.58/0.667/0.503 \\
        Zero-DCE \cite{guo2020zero} & 17.90/0.858/0.376 & 12.58/0.721/0.460 & 14.45/0.831/0.419 & 10.39/0.518/0.464 & 12.32/0.308/0.481 & 13.53/0.649/0.432 \\
        SCI \cite{ma2022toward} & 7.76/0.692/0.525 & 19.77/0.802/0.674 & 10.08/0.772/0.520 & 13.44/0.658/0.435 & 18.16/0.503/0.475 & 13.84/0.689/0.510 \\
        IAT \cite{cui2022you} & 14.46/0.705/0.386 & 18.70/0.780/0.665 & 17.88/0.829/0.547 & 13.65/0.616/0.528 & 13.87/0.317/0.536 & 15.71/0.649/0.532 \\
        RUAS \cite{liu2021retinex} & 17.89/0.869/0.336 & 12.59/0.721/0.459 & 17.05/0.808/0.575 & 10.39/0.519/0.463 & 12.33/0.308/0.479 & 14.05/0.645/0.462 \\
        Retinexformer \cite{cai2023retinexformer} & \underline{24.90}/\underline{0.891}/\underline{0.308} & 20.30/0.821/0.581 & 19.10/0.868/0.432 & 13.57/0.659/0.454 & 16.61/0.412/0.539 & 21.71/0.730/0.463 \\
        \midrule
        \multicolumn{7}{c}{Video enhancement methods} \\
        \midrule
        MBLLEN \cite{lv2018mbllen} & 22.39/0.877/0.353 & 23.59/0.788/0.559 & 19.99/0.836/0.542 & 14.09/0.636/0.525 & 13.17/0.501/0.555 & 18.65/0.728/0.507 \\
        LLVE \cite{zhang2021learning} & 19.97/0.848/0.393 & 15.17/0.764/0.610 & 18.17/0.855/0.465 & 13.84/0.638/0.492 & 15.35/0.287/0.577 & 16.50/0.678/0.507 \\
        \midrule
        \multicolumn{7}{c}{End-to-end enhancement methods} \\
        \midrule
        LLNeRF \cite{wang2023lighting}     & 24.05/0.849/0.351 & \underline{27.06}/\underline{0.863}/\underline{0.400} & \underline{26.17}/\underline{0.878}/0.416 & 20.17/\textbf{0.788}/\textbf{0.358} & 18.22/\textbf{0.643}/\textbf{0.384} & \underline{23.13}/\underline{0.804}/\underline{0.382} \\
        Aleth-NeRF \cite{cui2024aleth}   & 20.22/0.859/0.315 & 20.93/0.818/0.468 & 19.52/0.857/\underline{0.354} & \underline{20.46}/0.727/0.499 & \underline{18.24}/0.511/0.448 & 19.87/0 754/0.417 \\
        \pname & \textbf{30.03}/\textbf{0.918}/\textbf{0.274} & \textbf{29.27}/\textbf{0.879}/\textbf{0.345} & \textbf{30.00}/\textbf{0.900}/\textbf{0.380} & \textbf{23.82}/\underline{0.782}/\underline{0.451} & \textbf{18.26}/\underline{0.561}/\underline{0.439} &\textbf{26.28}/\textbf{0.808}/\textbf{0.379}  \\
        \bottomrule
    \end{tabular}
    } 
    \label{tab1}
\end{table*}

\begin{figure*}[h!]
    \centering
    \includegraphics[width=1\textwidth]{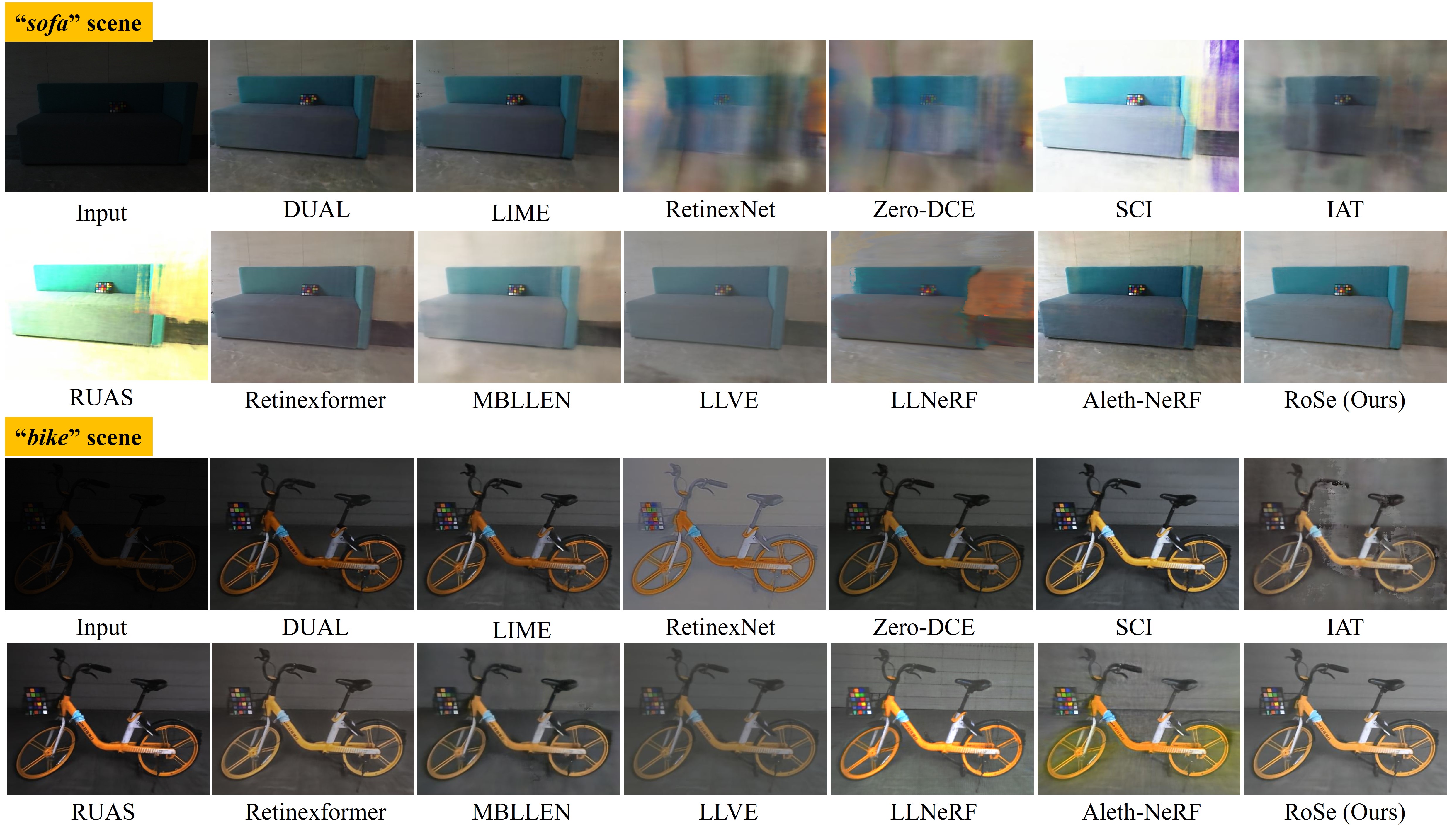}
    \vspace{-3mm}
    \caption{Normal-light novel view synthesis comparison in low-light conditions.}
    \label{fig:exp1}
\end{figure*}

As a result, the overall loss function for RetinexLight is defined as follows:
\begin{equation}
    \mathcal{L}=\mathcal{L}_{\text{MSE}}+ \lambda \mathcal{L}_{IC},
\end{equation}
where $\lambda$ is set to $1e^{-3}$.

\section{Experiments}

\noindent
{\bf Dataset.}
We conducted all experiments on the LOM dataset \cite{cui2024aleth}, which contains five scenes with 25–48 images captured under different exposure settings to simulate low-light and normal-light conditions. For dataset split, we used 3–5 images for testing, 1 image for validation, and the rest for training in each scene.

\noindent
{\bf Implementation details.}
During training, we used the Adam optimizer with an initial learning rate of $5e-4$ and cosine learning rate decay every 2500 iterations. The training batch size was 1024, and the total number of iterations was 75000. We employed hierarchical sampling with 64 coarse and 128 fine samples per ray, as in NeRF \cite{mildenhall2021nerf}. For fair comparisons, all methods were retrained on the LOM dataset using their default optimal settings.

\subsection{Main Results}

\begin{figure*}[h!]
    \centering
    \includegraphics[width=1\textwidth]{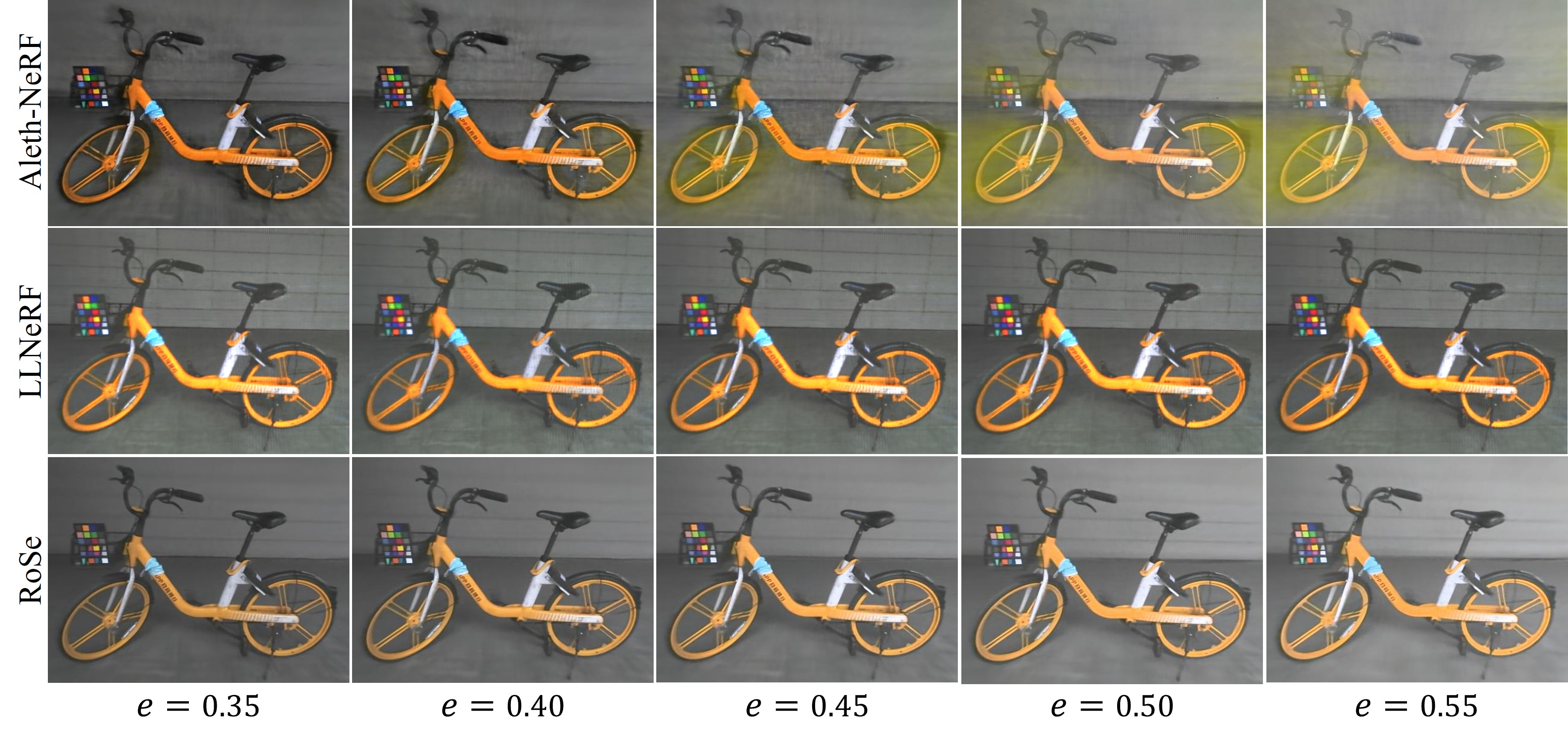}
    \caption{Ablation study on the effect of illumination level $e$ in illumination correction loss. \pname achieves robust illumination recovery across varying illumination levels without artifacts or color distortion.
    }
    \label{fig:ab-light}
\end{figure*}

\noindent
{\bf Quantitative comparison.}
We quantitatively compared \pname with state-of-the-art low-light enhancement and end-to-end methods using PSNR, SSIM, and LPIPS metrics across five scenes.
As a baseline, we first trained a vanilla NeRF under low-light conditions (\Cref{tab1}, first row). 

We then applied several traditional and advanced low-light enhancement methods to preprocess images before NeRF training, not but for postprocessing, as prior studies \cite{cui2024aleth} showed that postprocessing generally underperforms compared to preprocessing.
We also compared \pname with advanced end-to-end methods, LLNeRF and Aleth-NeRF.

As shown in \Cref{tab1}, traditional methods \cite{guo2016lime,zhang2019dual} performed poorly due to noise amplification during enhancement. Learning-based methods \cite{WeiWY018,guo2020zero,ma2022toward,cui2022you,liu2021retinex,cai2023retinexformer,lv2018mbllen,zhang2021learning} also failed to improve significantly due to color and structural distortions from multiview inconsistencies. Overall, end-to-end methods \cite{wang2023lighting,cui2024aleth} outperformed enhancement-based approaches, with our method achieving the best or second-best performance across all five scenes.

\begin{table*}[h!]
    \centering
    \caption{Ablation study on the order of low-rank denoising module (LRD) and illuminance transition MLP $F_i$. }
    \resizebox{1\textwidth}{!}{
    \begin{tabular}{lccccccc}
        \toprule
         \multirow{2}{*}{\textbf{Method}} & \textbf{“buu”} & \textbf{“chair”} & \textbf{“sofa”} & \textbf{“bike”} & \textbf{“shrub”} & \textbf{mean} \\
    \cline{2-7}
    & PSNR/SSIM/LPIPS & PSNR/SSIM/LPIPS & PSNR/SSIM/LPIPS & PSNR/SSIM/LPIPS & PSNR/SSIM/LPIPS & PSNR/SSIM/LPIPS \\
        \midrule
        MLP-first & 30.00/0.919/0.273 & 29.21/0.849/0.350 & 30.07/0.899/0.381 & 23.93/0.781/0.449 & 17.95/0.518/0.478 & 26.23/0.793/0.386 \\
        LRD-first & 30.03/0.918/0.274 & 29.27/0.879/0.345 & 30.00/0.900/0.380 & 23.82/0.782/0.451 & 18.26/0.561/0.439 & 26.28/0.808/0.379 \\
        \bottomrule
    \end{tabular}
    }
    \label{ab2}
\end{table*}

\noindent
{\bf Qualitative comparison.}
\Cref{fig:exp1} presents qualitative visualizations of the "sofa" and "bicycle" scenes from the LOM dataset. 
It is evident that low-light enhancement methods fail to recover reasonable scene brightness and structure when integrated with NeRF due to their view-inconsistent enhancement. 
Current state-of-the-art end-to-end models also show limitations: LLNeRF exhibits noticeable artifacts in both scenes due to conflicts introduced by its viewpoint-dependent enhancement.
Aleth-NeRF suffers from color distortions due to alterations in the properties of air particles.
In comparison, \pname generates more stable and vivid novel view images while better preserving the scene structure.

\subsection{Ablation Study}
\noindent
{\bf Effect of illumination level in $\mathcal{L}_{IC}$.}
\label{hyper}
The illumination level of the reconstructed target scene should not be fixed, as the precise level is unknown. 
We conducted ablation studies on the hyperparameter $e$ for illumination correction, comparing it to analogous parameters in Aleth-NeRF and LLNeRF.
As shown in \Cref{fig:ab-light}, Aleth-NeRF is highly sensitive to hyperparameter adjustments, leading to color distortions and diffusion artifacts. This instability stems from its method of modeling light attenuation by assigning transmittance values to air particles. When there is a large illumination gap, Aleth-NeRF misinterprets air as scene geometry. LLNeRF, on the other hand, lacks controllable illumination adjustment due to its view-inconsistent enhancement, which also results in artifacts.
In contrast, \pname demonstrates stable illumination adjustment capabilities across various hyperparameter settings without compromising structural integrity. 

\begin{table*}[h!]
    \centering
    \caption{Ablation study on the low-rank denoising (LRD) module. “w/o” and “w/” denote the absence and presence, respectively. “TV” refers to total variation loss \cite{kimmel2003variational}.}
    \vspace{-3mm}
    \resizebox{1\textwidth}{!}{
    \begin{tabular}{lccccccc}
        \toprule
         \multirow{2}{*}{\textbf{Method}} & \textbf{“buu”} & \textbf{“chair”} & \textbf{“sofa”} & \textbf{“bike”} & \textbf{“shrub”} & \textbf{mean} \\
    \cline{2-7}
    & PSNR/SSIM/LPIPS & PSNR/SSIM/LPIPS & PSNR/SSIM/LPIPS & PSNR/SSIM/LPIPS & PSNR/SSIM/LPIPS & PSNR/SSIM/LPIPS \\
        \midrule
        
        w/o LRD & 25.12/0.909/0.281 & 20.69/0.827/0.503 & 20.71/0.883/0.384 & 21.27/0.766/0.418 & 17.76/0.459/0.494 & 21.11/0.769/0.416\\
        w/ TV & 25.33/0.909/0.285 & 20.75/0.827/0.503 & 20.74/0.883/0.387 & 20.52/0.764/0.424 & 17.94/0.518/0.478 & 21.06/0.780/0.415 \\
        w/ LRD & 30.03/0.918/0.274 & 29.27/0.879/0.345 & 30.00/0.900/0.380 & 23.82/0.782/0.451 & 18.26/0.561/0.439 & 26.28/0.808/0.379 \\
        LRD+TV & 30.03/0.919/0.274 & 29.27/0.849/0.345 & 30.01/0.900/0.379 & 23.82/0.779/0.451 & 18.15/0.500/0.464 & 26.27/0.789/0.383 \\
        \bottomrule
    \end{tabular}
    }
    \label{ab3}
\end{table*}

\noindent
{\bf Order of LRD and illuminance transition MLP $F_i$.}
We conducted ablation studies on the ordering of the low-rank denoising module (LRD) and illuminance transition MLP $F_i$ to identify the optimal position of LRD. 
Specifically, we compared two configurations: LRD-first and MLP-first. 
As shown in \Cref{ab2}, while both configurations yield comparable performance, LRD-first performs better on PSNR, SSIM, and LPIPS metrics. 
We therefore adopt the LRD-first configuration in our final model.

\begin{figure}[h!]
    \centering
    \includegraphics[width=0.47\textwidth]{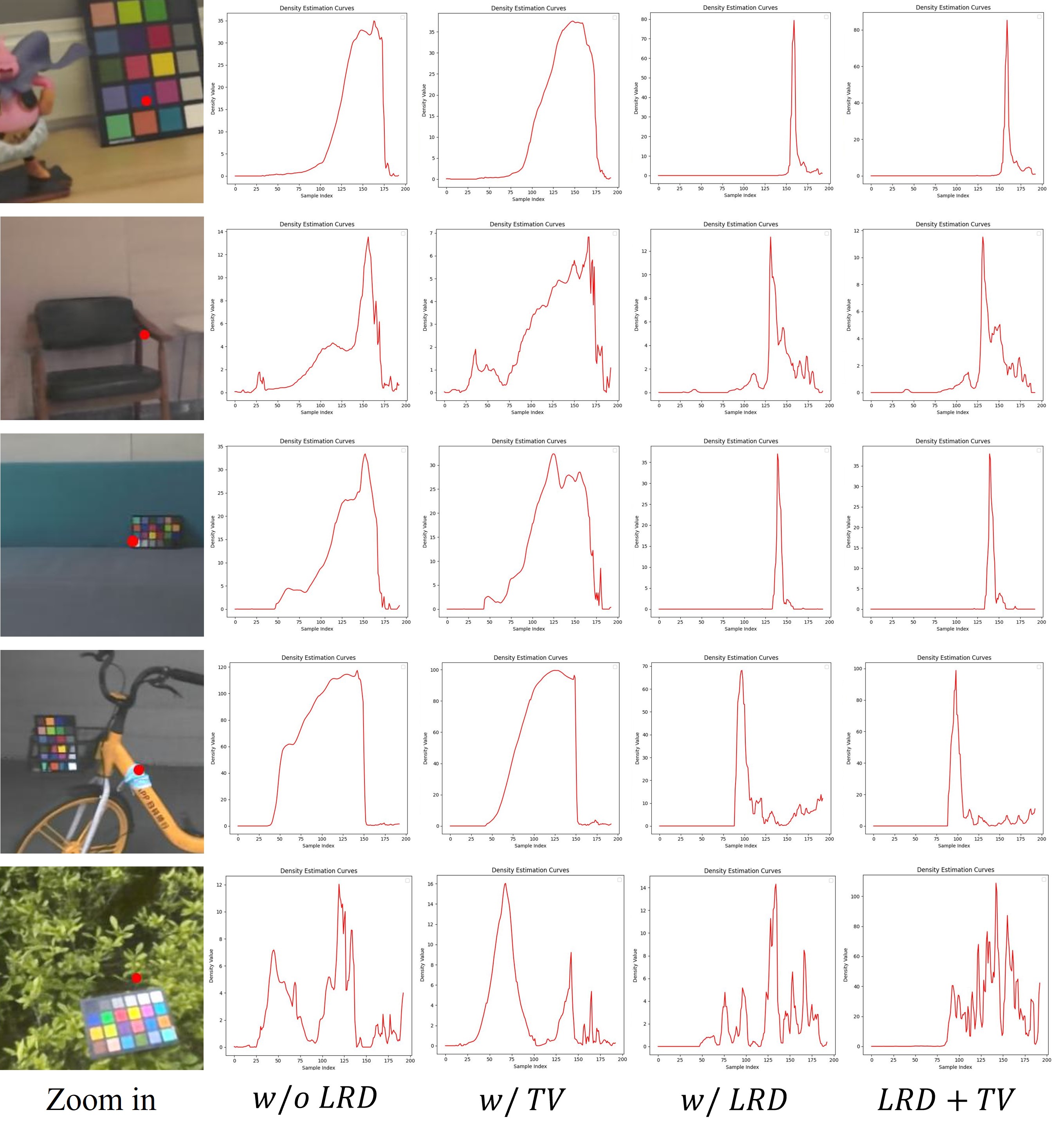}
    \caption{Density distribution of sampling points along the camera ray, with zoomed-in image pixels for better observation. “w/o” and “w/” denote the absence and presence, respectively. “TV” refers to total variation loss \cite{kimmel2003variational}.}
    \vspace{-3mm}

    \label{fig:ab-density}
\end{figure}

\noindent
{\bf Effect of low-rank denoising.}
The smoothness of illumination implies a low-rank structure, which we leverage via a low-rank denoising (LRD) module. To validate its denoising effect, we compare four settings: (1) without LRD (w/o LRD), (2) with total variation (TV) loss \cite{kimmel2003variational} (w/ TV) as an alternative, (3) with LRD (w/ LRD), and (4) LRD combined with TV loss (LRD+TV).

As shown in \Cref{fig:ab-density}, without LRD, noise causes incorrect object density estimation, leading to broader density distributions.
While TV loss provides some denoising effects, LRD achieves superior results with narrower distributions.
The combination of LRD and TV loss does not yield additional improvements, due to their overlapping roles.
Quantitative results in \Cref{ab3} confirm that LRD significantly improves performance, whereas TV loss provides limited benefit due to detail loss during denoising.
In complex scenes like “bike” and “shrub,” LRD’s gains are relatively modest, as these scenes require more complex illuminance transition modeling, which can be addressed by increasing sampling points.

\vspace{-3mm}
\section{Conclusions}
In this paper, we propose \pname, an unsupervised approach for robust low-light scene restoration. Unlike conventional Retinex-based decomposition and enhancement, we reformulate low-light restoration as an illuminance transition estimation problem and conceptualize this as a specialized neural radiance field. This leads to a concise dual-branch network consisting of a viewer-centered normal-light scene representation branch and a world-centered illuminance transition estimation branch, with a low-rank denoising module integrated to eliminate artifacts and improve structural integrity. Extensive experiments demonstrate that \pname outperforms current state-of-the-art methods in both visual quality and multi-view consistency.

RoSe currently relies on the Lambertian assumption \cite{smith1980lambertian}, and explicit modeling of specular reflection will be explored in the future to improve realism.
Moreover, \pname can also be integrated into efficient NeRF variants \cite{chen2022tensorf, muller2022instant} to accelerate training. However, this efficiency gain currently comes at the cost of quality degradation.
Addressing this trade-off -- as well as enabling generalization across scenes without per-scene training -- offers an important direction for future research.

\section*{Acknowledgments}
The work described in this paper was supported, in part, by the Innovation and Technology Fund (Projects: ITP/052/23TP \& ITP/004/24TP) and by the Research Institute for Intelligent Wearable Systems (Grant: CD95/ P0049355) and the Research Centre of Textiles for Future Fashion (Grants: BDVH/P0051330 \& BBFL/P0052601) of The Hong Kong Polytechnic University. 
It was also supported by the National Natural Science Foundation of China (Grant No. 62202241), Jiangsu Province Natural Science Foundation for Young Scholars (Grant No. BK20210586), NUPTSF (Grant No. NY221018), and Double-Innovation Doctor Program under Grant JSSCBS20220657.

{
    \small
    \bibliographystyle{ieeenat_fullname}
    \bibliography{main}
}

\end{document}